\documentclass[runningheads]{llncs}
\usepackage[T1]{fontenc}
\usepackage{graphicx}
\usepackage{amsmath}
\usepackage{algorithm}
\usepackage{algorithmic}
\usepackage{hyperref}
\usepackage{cite}
\begin{document}
\title{DRF: LLM-AGENT Dynamic Reputation Filtering Framework}
%
%
\author{Yuwei Lou\inst{1}\orcidID{0009-0002-5058-6401} \and
Hao Hu \inst{1}\orcidID{0009-0001-8277-9876} \and
Shaocong Ma\inst{1}\orcidID{0009-0007-4414-0303} \and
Zongfei Zhang\inst{2}\orcidID{0009-0000-6702-8807} \and
Liang Wang \inst{1} \orcidID{0000-0001-5444-748X} \and
Jidong Ge \inst{1} \orcidID{0000-0003-1773-0942} \and
Xianping Tao \inst{1}\orcidID{0000-0002-5536-3891}} 

\authorrunning{Y. Lou et al.}
%
\institute{State Key Laboratory for Novel Software Technology, Nanjing University, China
\email{\{yuweilou,mashaocong\}@smail.nju.edu.cn, \{myou,wl,gjd,txp\}@nju.edu.cn}\\
SCOT – Optimal Sourcing Systems, Amazon.com Services LLC, Washington, USA\\
\email{zongfe@amazon.com}}
\maketitle              
\begin{abstract}
With the evolution of generative AI, multi - agent systems leveraging large - language models(LLMs) have emerged as a powerful tool for complex tasks. However, these systems face challenges in quantifying agent performance and lack mechanisms to assess agent credibility. To address these issues, we introduce DRF, a dynamic reputation filtering framework. DRF constructs an interactive rating network to quantify agent performance, designs a reputation scoring mechanism to measure agent honesty and capability, and integrates an Upper Confidence Bound - based strategy to enhance agent selection efficiency. Experiments show that DRF significantly improves task completion quality and collaboration efficiency in logical reasoning and code - generation tasks, offering a new approach for multi - agent systems to handle large - scale tasks.
\keywords{LLM-Agent \and Team Optimization \and Generative AI.}
\end{abstract}

\section{Introduction}
In recent years, with the development of generative artificial intelligence, proxy-based artificial intelligence has achieved significant breakthroughs in practical applications, gradually evolving towards intelligent agent systems (In this paper, agent framework and agent systems are used interchangeably.) based on multiple Large Language Models (LLMs). 
These systems are capable of reasoning, learning, and collaboratively performing task actions, thereby promoting the transition of artificial intelligence from a single-model approach to a multi-agent collaborative mode. 
The emergence of research on multi-agent systems (MASs) based on large language models has made it possible for agents to collaborate efficiently in solving complex tasks and to support large-scale collective actions \cite{du2023improving, jiang2023llm, li2023camel, shinn2023reflexion, wang2307unleashing, wu2023autogen, zheng2023progressive}. 
The advent of these studies has shown that complex tasks can be solved not only by the intelligence of a single agent but also by the powerful potential of multi-agent collaboration in problem-solving.

Large language model (LLM) agents have shown excellent performance in various tasks, including affective human-computer interaction\cite{ren2025exploring}, code generation, and autonomous driving. Representative frameworks include Microsoft's open-source AutoGen and the popular CrewAI \cite{barbarroxa2024benchmarking}. These frameworks implement multi-agent systems through role specialization and task decomposition to automate complex objectives. They simulate virtual teams where each agent has a predefined role for accomplishing collective tasks.

As mentioned above, most current multi-agent frameworks use agent teams with different roles (e.g., software engineer, test engineer) to collaborate on tasks. However, these approaches have three key limitations: predefined role allocation is heavily human-expert dependent. there's an over-reliance on agent reliability without considering potential malicious interference or adversarial prompt injection, and there's a lack of agent capability differentiation, ignoring competency variations that may make some agents unsuitable for specific roles like software engineers.

We propose that an optimal multi-agent framework should form teams without specific task constraints, enabling agents to dynamically adapt to unknown task requirements. The framework also needs a robust mechanism to autonomously identify the most capable and efficient agents for task execution and systematically eliminate underperforming or malicious participants. To achieve these goals, this study addresses the challenge of building multi-agent teams in environments with malicious or low-quality agents. We develop a reinforcement learning framework to dynamically select high-performing agents. Our main contributions are threefold:

\begin{enumerate}
    \item 
    We propose an interactive rating network that dynamically assesses agent performance during task execution, enabling quantification of agent effectiveness.
    \item 
    We introduce a reputation iteration mechanism to rigorously evaluate agent reputation and capability, significantly mitigating task risks posed by low-efficiency agents.
    \item 
    We unify the rating network and the reputation iteration mechanism into an adaptive UCB selection architecture. This architecture shows excellent performance in enhancing task completion quality and cost-effectiveness, and is verified through real-world benchmark tests.
\end{enumerate}
\begin{table}[t]
\caption{Overview of related work.}\label{tab:Overview of related work}
\begin{tabular}{|p{4cm}|p{4cm}|p{4cm}|}
\hline
Research & Main idea & Reference \\
\hline
Prompt Engineering &  {Optimize the capability and accuracy of a single large model using prompt words.} & \cite{white2023prompt,zheng2023progressive} \\
Team Collaboration &  {Optimize team structure and strategies to enhance the capability and accuracy of large models.} & \cite{jiang2023llm, wu2023autogen, xiong2023examining, chen2023agentverse, liu2023dynamic, zheng2017truth} \\
\hline
\end{tabular}
\end{table}

\section{Related Work}
\label{sec:related work}
With the rapid advancement of artificial intelligence, generative AI has achieved significant breakthroughs in capabilities. Many generative AI systems, such as ChatGPT \cite{an2023correspondence}, Claude, and Deepseek \cite{liu2024deepseek}, have demonstrated transformative abilities in mathematical reasoning, code generation, and general language tasks. The emergence of large language models (LLMs) has given rise to the field of prompt engineering \cite{white2023prompt, zheng2023progressive}, which focuses on refining and optimizing input prompts to further unlock the potential of these models \cite{kojima2022large}. However, individual agents still face limitations in handling complex reasoning tasks, leading to the development of multi-agent systems (MASs) based on LLMs. Research on multi-agent systems has progressed as follows.
Although the optimization of multi-agent systems is a relatively recent area of interest, the study of optimizing human teams has a long history [17]. Current research includes Microsoft's Autogen [10], a framework that enables complex workflows through agent communication. It supports predefined agent types and employs a unique communication mechanism, utilizing an Orchestrator-Workers model to coordinate the system's operations. 

Xiong et al. \cite{xiong2023examining} proposed in their 2023 work a method where multiple agents engage in a ``tit-for-tat'' debate to express their arguments, thereby optimizing team performance through dialectical interaction. Similarly, Chen et al. \cite{chen2023agentverse} introduced the AgentVerse framework in 2023, which dynamically adjusts team composition and collaboration mechanisms based on task progress to enhance the performance of multi-agent systems. Liu Z et al. \cite{liu2023dynamic} proposed DyLAN, a dynamic feedforward network framework that selects agents through mutual evaluation among LLMs and deactivates underperforming agents. While their evaluation network inspired our research, we believe their early-stopping mechanism for task-participating agents may overlook the potential contributions of those agents. Therefore, in Section 4, we design and implement a k-layer scoring network that comprehensively evaluates and quantifies the contributions of all participating agents.

Reputation is an intrinsic attribute. Most of the previous research has focused on trust detection in the presence of Sybil attacks, but recent studies rarely assign reputation values to LLMs. Fortunately, Bouchiha M A et al. \cite{zheng2017truth} proposed a blockchain-based decentralized reputation system, which combines automated evaluation with human feedback to assign context-aware reputation scores that accurately reflect LLM behavior. However, their reliance on blockchain technology introduces additional overhead and time consumption due to smart contract operations. In this paper, we propose a more practical reputation scoring mechanism to help teams identify and prioritize agents with the highest reputation and capabilities.

Jiang et al. \cite{jiang2023llm} used additional LLMs in their work to rank the contributions of agents, thereby improving task accuracy and efficiency. Their approach of evaluating task progress to rank agents inspired our research. In Section 4, we detail how our multi-agent task scheduling strategy leverages the contribution values generated by the scoring network to influence agent reputation, subsequently altering agent selection criteria using an UCB approach.

\begin{figure}[htbp]
\includegraphics[width=\textwidth]{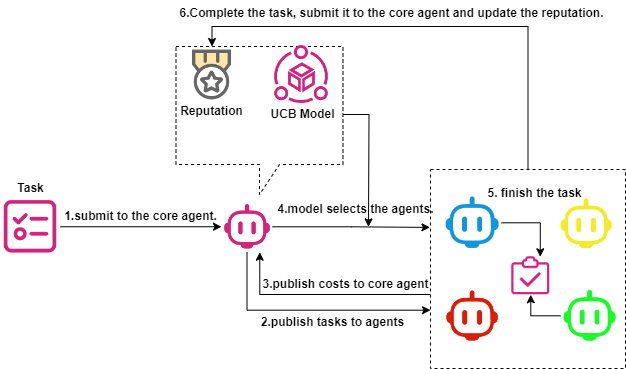}
\caption{The framework for DRF} \label{fig:The framework for DRF}
\end{figure}

\section{System Model and Problem Statement}
\subsection{System Modeling}
This section establishes the foundational modeling and conceptual definitions for the proposed multi-agent framework. As illustrated in Figure \ref{fig:The framework for DRF}, our framework introduces a novel agent team structure comprising a core agent and multiple task agents. The core agent is responsible for decision-making and control within the team, while the task agents primarily execute and evaluate tasks.

Definition 1 (Task, Round). When the core agent receives an input task, it evaluates the task's executability. A task is defined as a problem that can be resolved within a finite number of steps. Each task consists of multiple executable subtasks, and each subtask can be completed within a limited number of rounds. We denote the task set as $ T_{t}=\left\{S T_{t, 1}, S T_{t, 2}, S T_{t, 3}, \ldots S T_{t, m}\right\} $, where  represents the m-th subtask in the t-th round.

Definition 2 (Reputation). At the beginning of a task, the attributes of agents within the team are unknown. This implies that the team may consist of agents with varying capabilities, some of which may underperform or even be compromised by malicious prompt injections during task execution. Reputation is an intrinsic property of an agent, reflecting its trustworthiness and capability in completing tasks. However, reputation cannot be directly extracted from agents. To this end, we define a reputation score set for the agent team as $ R=\left\{r_{1}^{t}, r_{2}^{t}, \ldots, r_{i}^{t}\right\} $, which serves to quantify the scores of the agents.

Definition 3 (Accuracy) Accuracy measures the alignment between an agent's task completion and the actual task requirements. Higher accuracy indicates a smaller discrepancy between the agent's output and the task's expectations, signifying superior task performance. 

Definition 4 (Cost) In round t, each task agent submits a bid $ C_{i}^{t} $ to the core agent, representing its proposed cost for participating in the task(such as API calls). While agents submit bids in every round, it may not be selected by the core agent every time.

\subsection{Problem Formulation}
The agent team aims to select the agents with the lowest cost and highest efficiency to complete the task. Therefore, the main optimization objectives in this paper are as follows:

(1) Minimize the task execution cost, that is, the remuneration spent by the multi-agent team to solve the task should be the lowest. Suppose the agent team participating in solving the task in round $ t $ is $ P^{t} $, and the cost consumed by agent $ i $ in the participating team is $ C_{i}^{t} $. Then, the average cost consumed by the team in round $ t $ is as follows:
\begin{equation}
   \bar{C}^{t}=\frac{\sum_{i=1}^{\left|P^{t}\right|} C_{i}^{t}}{\left|P^{t}\right|}
\end{equation}

(2) The highest average payoff. This paper aims to maximize the team's payoff for each task, meaning that every time the task is executed, the team can achieve the degree of meeting the task requirements to the greatest extent, i.e., the highest accuracy. We use $ A_{i}^{t} $ to represent the accuracy of the task completed by agent $ i $ in round $ t $, which can be calculated using the following equation:
\begin{equation}
    A_{i}^{t}=\frac{\left|w_{0}^{t}-w_{i}^{t}\right|}{w_{0}^{t}}
\end{equation}

In Equation (2), $w_{i}^{t}$ represents the score of agent $i$ in round $ t $, while denotes the score threshold within the same paper. Detailed definitions of these symbols will be provided in Section 4. Then, the average accuracy of the agent team participating in the task in round $ t $ is:
\begin{equation}
    \bar{A}^{t}=\frac{\sum_{i=1}^{\left|P^{t}\right|} A_{i}^{t}}{\left|P^{t}\right|}
\end{equation}
In summary, the main research objectives of this paper can be obtained as follows:
\begin{equation}
    \left\{\begin{array}{l}\operatorname{Min}(\bar{C}^{t})=\operatorname{Min}\left(\frac{\sum_{i=1}^{\left|P^{t}\right|} C_{i}^{t}}{\left|P^{t}\right|}\right) \\ \operatorname{Max}(\bar{A}^{t})=\operatorname{Max}\left(\frac{\sum_{i=1}^{\left|P^{t}\right|} A_{i}^{t}}{\left|P^{t}\right|}\right)\end{array}\right.
\end{equation}
\textbf{Subject to:}
\begin{equation}
    \sum_{t} \sum_{N} c_{i}^{t} \leq \emptyset, P^{t} \leq K
\end{equation}
The above $ \phi $ represents the total budget sent to the agent team before the task starts, $ P^{t} $ is the agent team participating in solving the task in round $t$, and $K$ is the maximum number of agents selected in each round.
\begin{figure}[t]
\includegraphics[width=\textwidth]{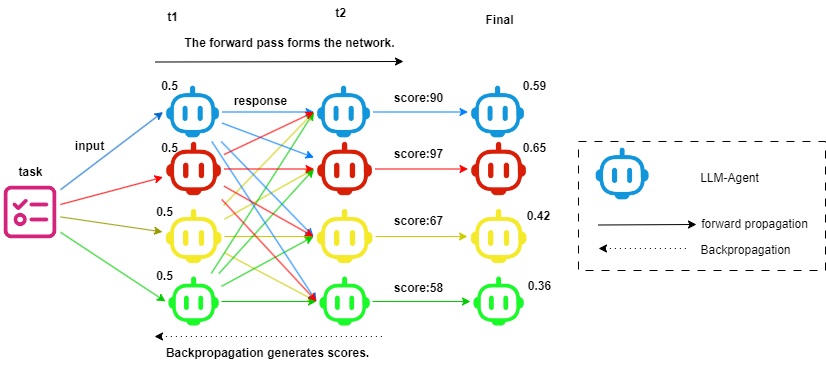}
\caption{Network Construction and Rating Process} \label{fig:Network Construction and Rating Process}
\end{figure}
\section{Model Construction}
This section provides a detailed description of the model construction. It aims to build a rating network for LLM agents to quantify their execution performance in tasks. A reputation iteration mechanism is also constructed to dynamically distinguish the reputation and capability of different agents. Additionally, a task scheduling strategy is developed to prioritize the utilization of agents with high reputation and low cost. This ultimately enables efficient task execution by multiple LLM agents.
\subsection{LLM-Agent Rating Network}
When a task is published within a multi-agent team, effectively measuring the performance of agents in task execution is a challenging problem. Inspired by the construction of neural networks and the work of DyLAN \cite{liu2023dynamic}, we propose a multi-agent rating network to quantify the task execution performance within a multi-agent team.
In this paper, the construction of the rating network system is divided into two stages: forward pass for network construction and backward pass for obtaining ratings. When a specific task is published to the multi-agent team, agents decide whether to participate in the task based on their own willingness and capabilities. This leads to the following model construction:

\textbf{Node}: In the rating network, a node represents an agent executing the task during a specific time interval.The agent's input is divided into two categories: task context information requiring a solution and solutions received from other agents in the previous time step. Correspondingly, the output is also divided into two categories: the solution to the input task and the ratings for the solutions received in the previous time step.  In other words, the $i$-th agent at time $t$ can be defined as a function $f_{i}^{t}$, with the prompt information denoted as $p_{i}$,The specific role of this function will be described in the following text.

\textbf{Edge}: In the rating network, an edge represents the relationship between each pair of agents. There are two types of relationships: one is the rating given by an agent to the solutions proposed by other agents in the previous time step, and the other is the solution proposed by the agent for the task in the current time step.

\textbf{Forward Pass for Network Construction:} 
Each task consists of multiple time rounds. In each round, an agent generates a solution based on the task and passes it to other agents. If there were solutions proposed by other agents in the previous round, the agent will receive and evaluate these solutions. The function $ f_{i}^{t}(p_{i}, S T_{t, m},responses) $  represents two scenarios in the network construction:if "responses" is empty, the agent is generating a solution; otherwise, it is evaluating one. Inspired by Reflexion\cite{shinn2023reflexion}, which demonstrates that using LLM as an evaluator can effectively overcome the limitations of traditional metrics in natural-language tasks, we adopt the same approach for all evaluating in this paper. And during task execution, as each agent may have potential value, no agent is eliminated during the network construction process. Following the above procedure, a $k$-layer rating network can be formed after $k$ rounds, as shown in Figure \ref{fig:Network Construction and Rating Process}.

\textbf{Backward Pass for Score Calculation:} 
After forming a $k$-layer rating network, backpropagation is performed to evaluate each agent's task contribution. Based on the forward pass, the score that agent $i$ receives from agent $j$ in round $t$ of the task is $w_{j,i}^{t} = f_{i}^{t}(p_{i}, S T_{t, m},responses)$ (note that agent $i$ and $j$ are different, meaning agents cannot rate their own solutions).so we can obtain the rating set for agent $i$ in round $t$ as $ \left[w_{1, i}^{t}, w_{2, i}^{t}, w_{3, i}^{t}, \ldots, w_{j, i}^{t}\right] $. We can also obtain the set of reputation values for all agents participating in the task as $ \left[r_{1}^{t-1}, r_{2}^{t-1}, \ldots, r_{j}^{t-1}\right] $,where $r_{j}^{t-1}$ represents the reputation value of agent $j$ in round $t-1$. For detailed descriptions and calculations, see Section 4.2.Then, we can calculate the score of the agent in round $t$ using Equation (6).
\begin{equation}
    w_{i}^{t}=\sum_{j=0}^{| P^t |} w_{j, i}^{t}*\varphi_j ,\varphi_j = \frac{e^{r_{j}^{t-1}}}{\sum_{k = 1}^{| P^t |}e^{r_{k}^{t-1}}}
\end{equation}

Here, $ w_{i}^{t}$ denotes the score of agent $i$ in round $t$, while $ \varphi_j $ represents the significance coefficient of the reputation of agent $j$ within the set of agents participating in the task $P^t$.This implies that evaluations from agents with higher reputation are typically more valuable, which makes the rating network more rational.

\begin{algorithm}[t]
\caption{DRF-Reputation Algorithm}
\begin{algorithmic}[1]
\REQUIRE reputation set \( R \), parameter \( w_o \), the coefficient of reputation increment \( \delta \), the penalty coefficient \( \beta \), selected agents set \( P^t \)
\ENSURE reputation set \( R \)
    \FOR{agent in \( P^t \)}
        \STATE Provide a Task Plan
        \STATE Deliver to the remaining agents for evaluation.
        \STATE Join the forward pass to form the network.
        \IF{agent obtains all the evaluations}
            \STATE Initiate backpropagation
            \STATE calculate \( w_i^t \) by using Eq.(6)
        \ENDIF
        \IF{\( w_o > w_i^t \)}
            \STATE update \( R[\text{worker}] \) using Eq.(8)
        \ELSE
            \STATE update \( R[\text{worker}] \) using Eq.(7)
        \ENDIF
    \ENDFOR
    \RETURN \( R \)
\end{algorithmic}
\end{algorithm}

\subsection{LLM-Agent Reputation Iteration Mechanism}
When multiple agents begin to execute tasks, they will form a rating network based on the content described in Section 4.1. After completing one round of rating, each agent will obtain its task score. Since there may be some agents with lower capabilities or those engaging in malicious interference within the group completing the task, it is necessary to assign reputation scores to the agents. This helps decision-makers better select agents from the group.

For each agent $i$ participating in the task in round $t$, we assign a reputation value $r_{i }^{t}$ to measure its credibility (reputation). We consider $w_{i}^{t}$ as a performance metric of an agent, which is determined by its own capabilities and characteristics (such as honesty). A more honest agent, that is, one with a higher reputation, is expected to complete the task with better quality, meaning a larger $w_{i}^{t}$ value. Therefore, we need to design a mechanism to better evaluate the reputation of agents.\\
\textbf{Reputation Increment Mechanism:}\\
Each time a task rating network is formed, we can obtain the task score $w_{i}^{t}$ for each agent. During the task execution, there is a specific task threshold $w_{o}$ for each task, which is used to measure whether the task solution provided by the agent can be well accomplished. The threshold is generally set according to task requirements and empirical experience. If $ w_{o} \leq w_{i}^{t} $, it indicates that agent $i$ has performed well in that task time interval and its reputation should be increased, as shown in Equation (7).
\begin{equation}
    r_{i}^{t}=r_{i}^{t-1}+w_{i}^{t} *\left(1-r_{i}^{t-1}\right) * \partial
\end{equation}

Here, $r_{i}^{t}$ is the reputation value of agent $i$ in round $t$, $r_{i}^{t-1}$ is the reputation value of agent $i$ in round $t-1$, $w_i^t$ is the task score of agent $i$ in round $t$, and $\alpha$ is the coefficient of reputation increment.\\
\textbf{Reputation Decay Mechanism:}\\
Similar to the reputation increment mechanism, when $ w_{o}>w_{i}^{t} $, it indicates that agent $i$ has performed poorly in that task time interval, likely due to malicious interference or inherently being a subpar agent. Therefore, we need to use Equation (8) to decrease the reputation of the agent.
\begin{equation}
    r_{i}^{t}=r_{i}^{t-1}-\left(w_{i}^{t} * r_{i}^{t-1}\right) * \beta
\end{equation}

Here, $r_{i}^{t}$ is the reputation value of agent $i$ in round $t$, $r_{i}^{t-1}$ is the reputation value of agent $i$ in round $t-1$, $w_{i}^{t}$ is the task score of agent $i$ in round $t$, and $\beta$ is the penalty coefficient for reputation. According to Equation (12), we can see that the lower the task score $w_{i}^{t}$, the greater the decline in reputation, allowing for quicker identification of malicious or underperforming agents.

\textbf{Algorithm 1} illustrates the process of the reputation iteration mechanism we employ, incorporating a multi-agent rating network. Here, $R$ denotes the set of agent reputations, and $P^t$ represents the set of agents participating in the task.

\subsection{LLM-Agent Task Scheduling Strategy}
For each agent team participating in a task, we can effectively rate and monitor the agents through the multi - agent rating network in Section 4.1, and evaluate agent attributes via the multi - agent reputation iteration mechanism in Section 4.2. But in real - world scenarios, we initially don't know which agent has the highest reputation or performs best in the task. So, we need a strategy to gradually explore and filter out more competent agents from this unknown situation.

The UCB-based Multi-Armed Bandit algorithm \cite{lou2024mab, garivier2011kl} is a reinforcement learning method that balances exploration and exploitation. In our model, each agent is like an arm of the MAB. Selecting an agent is analogous to pulling an arm. The outcome of each pull depends only on the current arm chosen, not on previously selected arms or past results. This meets the Markov property requirement. The algorithm helps choose between agents with the highest current task payoff (exploitation) and those that might offer higher future payoffs (exploration).

At the outset of the task, we lack information on the agents' attributes and reputation values. Hence, it is imperative to swiftly identify agents with high reputation values. As the algorithm progresses, we incrementally uncover the reputation values of some agents through continuous exploration and exploitation. Consequently, it becomes crucial to consider the overall payoff of agent recruitment. In this paper, the overall payoff incorporates multiple factors, including reputation and cost (with cost denoting the expense of agent invocation). To address these two scenarios, we have redesigned the UCB algorithm to facilitate the selection of efficient agents for task execution by the agent team, as illustrated in Equation (9).Here, $S_i^t$ is the selection basis value for agent $i$ in round $t$, $r_i^{t-1}$ is the reputation value of agent $i$ in round $t-1$, $\gamma$ is a positive parameter that allows multiple possibilities for the influence of $x_i^{t-1}$, and $n_i^{t-1}$ is the number of times agent $i$ was selected in round $t-1$,$c_i^t$ represents the cost incurred by agent $i$ in executing the task in round $t$.$\delta$ represents a weight coefficient that adjusts the proportion between reputation and cost.
\begin{equation}
    S_{i}^{t}=\delta * r_{i}^{t-1} + (1 - \delta)c_i^t +x_{i}^{t-1}, x_{i}^{t-1}=\sqrt{\frac{\gamma * \ln \left(\sum_{j=1}^{N} n_{j}^{t-1}\right)}{n_{i}^{t-1}}}
\end{equation}
Algorithm 2 provides a detailed illustration of the improved UCB algorithm integrated with the rating network and reputation updates in our DRF framework. During the execution of the UCB algorithm, the selection basis value $S_i^t$ is influenced by both the rating network and the changes in reputation.Here, $R_0$ denotes the upper bound of reputation. Agents exceeding this value are deemed trustworthy and are exempted from further reputation checks.
\begin{algorithm}[t]
\caption{DRF Selection Algorithm}
\begin{algorithmic}[1]
\REQUIRE The threshold of reputation \( R_0 \), reputation set \( R \), costs set \( C \), The total budget \( \phi \), parameters \( \gamma \), \( \beta \), \( \sigma \)
\STATE \( t = 1, init \ P^t \)
\WHILE{ \( \phi > 0 \) }
    \STATE obtain agents that want to participate in the task.
    \FOR{agent in agents}
        \IF{ \( R[\text{agent}] \geq R_0 \) }
            \STATE add the agent to \( P^t \).
        \ELSE
            \STATE calculate \( S_i^t \) by using Eq.(9) with a custom weight.
        \ENDIF
    \ENDFOR
    \STATE select agents by \( S_i^t \) and get \( P^t \).
    \FOR{agent in \( P^t \)}
        \STATE let agent do tasks and get a rating network.
        \STATE calculate the agent's \( w_i^t \).
        \STATE use Reputation Iteration Mechanism to adjust the agent's Reputation.
        \STATE perform payment for the agent.
        \STATE calculate the payoff.
    \ENDFOR
    \STATE \( t = t + 1 \)
\ENDWHILE
\end{algorithmic}
\end{algorithm}
\begin{table}[t]
\caption{LLM Prompt Setting.}\label{tab:LLM Prompt Setting}
\begin{tabular}{|p{4cm}|p{8cm}|}
\hline
Role &  Prompt \\
\hline
Executor &   You are a task-solving assistant capable of completing tasks based on the user-specified ability level (low, medium, high). The ability level determines the precision and complexity of your task completion:Low ability: Your answers are very simple, mostly incorrect, with a very low probability of being correct.Medium ability: Your answers are average, generating imprecise solutions that are not detailed or clear.High ability: Your answers are very precise and comprehensive, capable of handling complex tasks and providing optimal solutions.User Input:Please specify your ability level: {ability}.Please describe the task you need to complete: {task}.Output:Based on the ability level specified by the user, strictly generate a task solution that aligns with the description of your current ability level.  \\
\hline
Evaluator & You are a strict task evaluation assistant capable of assessing the answers provided by users to task solutions and assigning scores:If the task answer is very precise and comprehensive, capable of handling complex tasks and providing optimal solutions, score between 80-100;If the task answer is usable but not detailed or rigorous enough, score between 60-80;If the task answer is imprecise or contains errors, score below 60.User Input:Task solution answer:{answer}.Output: (Note that when outputting in markdown, the mask field should not be bolded; it should remain as plain text. For example:)mask: 85 ,des: The evaluation of the task \\
\hline
\end{tabular}
\end{table}

\begin{table}[t]
\caption{The experimental results on the HumanEval dataset. We conduct multiple tests and report the average results.} \label{tab:The experimental results on the HumanEval dataset}
\begin{tabular}{|p{2cm}|p{1.4cm}|p{1.6cm}|p{1.6cm}|p{1.6cm}|p{1.3cm}|p{1.3cm}|p{1.3cm}|}
\hline
Agent System & Select agent & Pass@1(6) & Pass@1(12) & Pass@1(18) &  Cost(6) & Cost(12) & Cost(18) \\
\hline
CodeT &  {\itshape --} & 64.5 & 64.3 & 65.8 & -- & -- & --\\
DyLAN &  {\itshape 3} & 80.2 & 83.1 & 88.3 & 0.84 & 0.86 & 0.81\\
Reflexion & {\itshape3} & 74.2 & 80.6 & 86.5 & 0.86 & 0.90 & 0.88\\
DRF(Ours) & {\itshape 3} & 84.3  & 86.5 & 92.9 & 0.76 & 0.74 & 0.71\\
\hline
\end{tabular}
\end{table}

\begin{table}[t]
\caption{The experimental results on the BigBench dataset.We conduct multiple tests and report the average accuracy.}\label{tab:The experimental results on the BigBench dataset}
\begin{tabular}{|p{2cm}|p{1.4cm}|p{1.5cm}|p{1.5cm}|p{1.5cm}|p{1.3cm}|p{1.3cm}|p{1.3cm}|}
\hline
Agent System & Select agent & Result(6) & Result(12)& Result(18) & Cost(6) & Cost(12) & Cost(18)\\
\hline
LLM-Debate &  {\itshape 5} & 59.4 & 61.6 & 63.4 & 0.89 & 0.88 & 0.91\\
DyLAN &  {\itshape 5}     & 62.5 & 64.4 & 66.2 & 0.84 & 0.86 & 0.83\\
Reflexion & {\itshape 5} & 53.1  & 58.2 & 60.3 & 0.86 & 0.89 & 0.87 \\
DRF(Ours) & {\itshape 5} & 64.6  & 65.9 & 70.5 & 0.79 & 0.81 & 0.75\\
\hline
\end{tabular}
\end{table}
\section{Experiment}
This section concentrates on the experimental settings and results. To demonstrate the efficiency and superiority of our proposed LLM - based agent systems, we conduct experiments using real - world datasets. We compare DRF with other LLM- based agent systems to assess its performance. The experiments are analyzed in terms of experimental preparation, performance experiments, and comparative experiments.\\
\subsection{Experimental Preparation}
Each of our agents consists of two LLM models, Executor(for task execution) and Evaluator(for task evaluation), both driven by the DeepSeek-R1 model. To simulate different agent capabilities, we input a meta-prompt to each model with a capability description field. ``Low'' indicates low capability or a malicious agent (which generates random or wrong solutions, similar to low capability). ``Medium'' means the agent may solve tasks correctly or incorrectly. ``High'' indicates the agent can efficiently complete tasks. We set three LLM agent combinations: 6-12-18, with 30\% low/high-capability agents and the rest medium. Hyperparameters $\alpha$ and $\beta$ are typically set based on machine learning experience, with a default value of 0.1. $\gamma$ is set to 2 following UCB algorithm conventions. The reputation threshold is set to 0.9 in this paper, and the initial reputation of each agent is set to 0.5. In previous studies, the cost of LLM agents was rarely discussed. Recently, DyLAN \cite{liu2023dynamic} measured it by API calls. To better simulate the cost differences among agents (which vary in reality), we used a uniform distribution to generate per-task costs for each agent, mapping all costs to (0,1).

We use two datasets for our experiments. For code tasks, we use the Humaneval dataset (Chen et al. \cite{garivier2011kl}), which is designed for code completion, to test the capabilities of different LLM agent systems and DRF. For logical reasoning tests, we use the logic grid puzzles task from the BigBench dataset (Srivastava et al., 2022 \cite{srivastava2022beyond}) to further compare these agent systems. We collect prompts from various LLM agent systems and have DeepSeek-R1 conduct further summarization and analysis. The prompts listed in Table \ref{tab:LLM Prompt Setting} are derived and used in all subsequent experiments.
\begin{figure}[t]
\includegraphics[width=\textwidth]{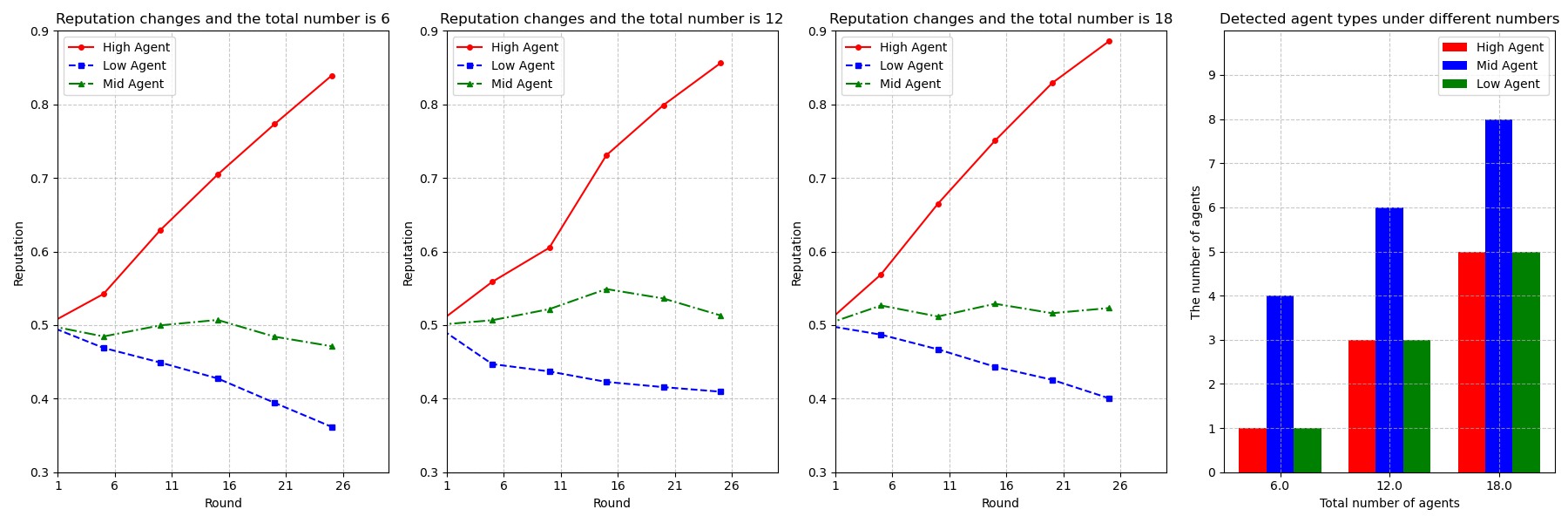}
\caption{Changes in the reputation of agents} \label{fig:Changes in the reputation of agents}
\end{figure}
\subsection{ Performance Experiments}
We conducted experiments on the Humaneval dataset, randomly sampled 30 test cases for experimentation, with each test case representing a round in the DRF framework. We set $\delta$ to 1 in DRF, focusing solely on reputation, and adjusted the temperature hyperparameter of the LLM to 0.2 to ensure coherent output. The results are shown in Figure \ref{fig:Changes in the reputation of agents}. It can be seen that DRF can increase the reputation of high - capability agents and decrease that of low - capability ones within limited rounds. The fluctuating reputation of mid - capability agents is due to their inconsistent task performance, leading to variable scores and reputation changes. Moreover, DRF variations with different numbers of agents can effectively detect the preset numbers of high, mid, and low - capability agents, confirming the effectiveness and feasibility of DRF.

\subsection{Comparative Experiment Performance}
The experiments have demonstrated that DRF can effectively identify high - reputation, high - quality LLM - Agents in a team during task execution through a rating network and reputation iteration mechanism. In the comparative experiments, we benchmark DRF against mainstream LLM agent frameworks to evaluate its efficiency.

For code tasks, we compare DRF with three LLM agent frameworks—DyLAN \cite{liu2023dynamic}, CodeT \cite{zheng2017truth}, and Reflexion \cite{shinn2023reflexion}. Experiments are conducted on the Humaneval dataset, with performance measured using the pass@1 metric. Agents for each framework are selected from Section 5.1 to ensure uniform external conditions. In each experiment, three agents are selected from the agent team to participate in the task, and the agent temperature is set to 0.2. In DRF, \(\delta\) is set to 0.7. For Reflexion, three agents are assigned to the roles of executor, evaluator, and refresher. In DyLAN, three agents perform the task and form a dynamic early-stopping network.

For logical reasoning tests, CodeT is no longer effective. We selected the more suitable deliberative multi-agent system LLM-Debate \cite{srivastava2022beyond} as a replacement. DyLAN, Reflexion, and DRF also participated in the task. In this experiment, we set the number of agents to select from the agent team at 5(Logical reasoning tasks are complex and require increasing the number of agents.), with the agent temperature at 0.8 (to encourage more diverse responses in reasoning tasks based on prompts), while other framework settings remained similar to those in the code task. We report the accuracy.

Table \ref{tab:The experimental results on the HumanEval dataset} shows the experimental results of four LLM agent systems with different numbers of agents. Pass@1(6) indicates the Pass@1 metric when there are six agents in total."Select agent" is the number of agents chosen from the total to participate in the task. As the total number of agents increases, the performance of DyLAN, Reflexion, and DRF improves. This is because a larger total number of agents leads to more high-reputation agents, increasing the likelihood of their participation and thus enhancing performance. DRF consistently outperforms the other systems across all metrics. It effectively filters agents based on reputation and prioritizes high-reputation, low-cost agents for subsequent tasks. Although DyLAN's early-stopping mechanism can identify low-reputation agents, it cannot exclude them in subsequent rounds. Reflexion, being more dependent on agent capability and lacking a filtering mechanism for low-reputation agents, underperforms DRF and DyLAN in experiments.

Table \ref{tab:The experimental results on the BigBench dataset} presents the experimental results for logical reasoning tests. The result(6) column shows the accuracy with six agents. DRF outperforms other LLM agent systems in terms of both performance and average cost. We attribute DRF's performance edge to its integrated rating network and reputation iteration mechanism, which facilitates the identification of high-reputation agents. Additionally, the task scheduling strategy effectively utilizes these high-reputation, low-cost agents and prioritizes their selection for subsequent tasks.

\section{CONCLUSION}
We introduce DRF, an LLM agent framework combining an interactive rating network, a reputation iteration mechanism, and a UCB selection strategy. It enhances collaboration efficiency and task completion quality in LLM-based multi-agent framework. Unlike conventional systems, DRF uses the UCB reinforcement learning method to leverage the reputation attribute of LLM agents.  Extensive real - dataset experiments show DRF excels in logical reasoning and code generation, outperforming other current agent frameworks under the same conditions.

Future work will focus on two directions. First, we'll explore using advanced reinforcement learning algorithms, such as DQN, in DRF to handle more complex and diverse tasks. Second, we'll investigate how to enhance certain LLMs with an experience - augmented reputation model, providing more LLM agent options for complex tasks.

\bibliographystyle{splncs04}
\bibliography{paper}

\end{document}